\documentclass[10pt,twocolumn,letterpaper]{article}

\usepackage{icb}
\usepackage{times}
\usepackage{epsfig}
\usepackage{graphicx}
\usepackage{amsmath}
\usepackage{amssymb}
\usepackage[table]{xcolor}
\usepackage{tabularx}
\usepackage[flushleft]{threeparttable} 
\usepackage{booktabs,caption}
\usepackage{fancyhdr}
\usepackage[caption=false,font=footnotesize]{subfig}
\usepackage{xpatch}
\usepackage[title]{appendix}
\usepackage{multirow}

\usepackage[pagebackref=true,breaklinks=true,letterpaper=true,colorlinks,bookmarks=false]{hyperref}

\hypersetup{
  colorlinks, linkcolor=red
}
\makeatletter
\g@addto@macro{\UrlBreaks}{\UrlOrds}
\makeatother

\definecolor{Gray}{gray}{0.90}
\newcolumntype{a}{>{\columncolor{Gray}}c}

\makeatletter
\xpatchcmd{\paragraph}{3.25ex \@plus1ex \@minus.2ex}{3pt plus 1pt minus 1pt}{\typeout{success!}}{\typeout{failure!}}
\makeatother

\newcolumntype{Y}{>{\centering\arraybackslash}X}
\newcolumntype{s}{>{\hsize=.3\hsize}X}
\newcolumntype{b}{>{\hsize=1.4\hsize}X}

\DeclareRobustCommand{\eg} {\textit{e}.\textit{g}.\@\xspace}
\DeclareRobustCommand{\ie}{\textit{i}.\textit{e}.\@\xspace}
\DeclareRobustCommand{\etal}{\textit{et al.}\@\xspace}

\icbfinalcopy 

\ificbfinal\fi
\begin{document}

\title{Actions Speak Louder Than (Pass)words:\\  Passive Authentication of Smartphone\thanks{Smartphones are multi-purpose mobile computing devices with strong hardware capabilities and extensive mobile operating systems, facilitating wide internet and multimedia functionalities~\cite{def_smartphone}.}~~Users via Deep Temporal Features}

\author{Debayan Deb, Arun Ross, Anil K. Jain\\
Michigan State University\\
East Lansing, MI, USA\\
{\tt\small{\{debdebay, rossarun, jain\}@cse.msu.edu}}
\and
Kwaku Prakah-Asante, K. Venkatesh Prasad\\
Ford Motor Company\\
Dearborn, MI, USA\\
{\tt\small{\{kprakaha, kprasad\}@ford.com}}}

\maketitle
\thispagestyle{empty}

\begin{abstract}
Prevailing user authentication schemes on smartphones rely on explicit user interaction, where a user types in a passcode or presents a biometric cue such as face, fingerprint, or iris. In addition to being cumbersome and obtrusive to the users, such authentication mechanisms pose security and privacy concerns. Passive authentication systems can tackle these challenges by frequently and unobtrusively monitoring the user's interaction with the device. In this paper, we propose a Siamese Long Short-Term Memory network architecture for passive authentication, where users can be verified without requiring any explicit authentication step. We acquired a dataset comprising of measurements from 30 smartphone sensor modalities for 37 users. We evaluate our approach on 8 dominant modalities, namely, keystroke dynamics, GPS location, accelerometer, gyroscope, magnetometer, linear accelerometer, gravity, and rotation sensors. Experimental results find that, within 3 seconds, a genuine user can be correctly verified 97.15\% of the time at a false accept rate of 0.1\%. \end{abstract}

\section{Introduction}
The Digital Age has ushered in a large number of devices that store and generate information. Among these devices, smartphones are the most widely used~\cite{smartphone_most_common}.  An interesting phenomenon has been observed in the smartphone age, where users appreciate the convenience of services at their fingertips and implicitly store more and more valuable and private data,~\eg, banking and payment details, health records, etc. This has inadvertently sparked a community of hackers that dedicate their time and effort to gain access to smartphones in order to steal sensitive data~\cite{steal_info}. Therefore, securing access to mobile devices by authenticating users is of utmost importance.

\begin{figure}[!t]
\centering
\includegraphics[width=0.65\linewidth]{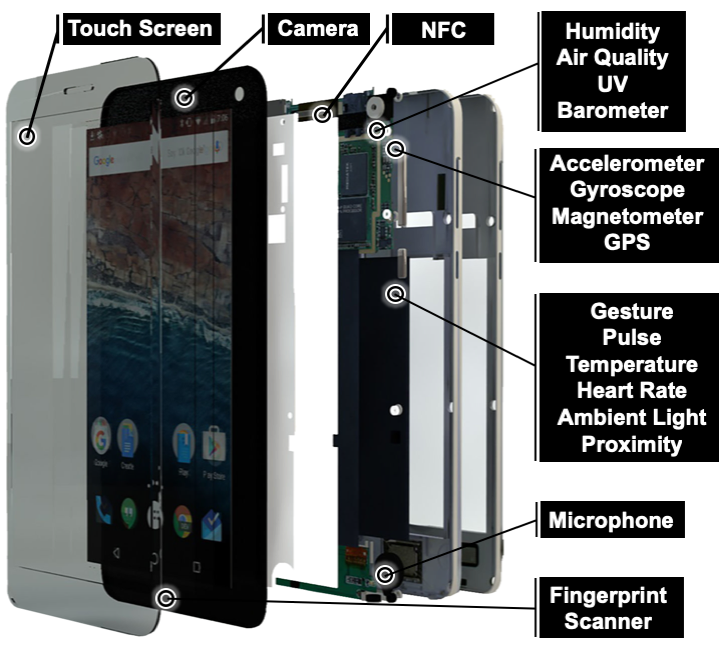}
\caption{Authentication on smartphones by exploiting sensorial data has become an active field of research due to the growing number of available sensors in  smartphones.}
\label{fig:sensors}
\end{figure}


Current authentication schemes on mobile platforms require explicit user interaction with the device, referred to as ~\emph{explicit authentication}, in order to gain access to it. The entry point for device access is typically a passcode or a biometric cue such as face, fingerprints or iris~\cite{delta_ID}. Passwords and PINs have long been viewed as the pinnacle of securing information and controlling access to mobile devices. However, these knowledge-based authentication schemes are prone to social engineering hacks, guessing and over-the-shoulder attacks~\cite{phone_guess}. With recent advances in technology, smartphones are getting better at authenticating users by learning their biological traits, such as face, fingerprint, or iris, which are believed to be unique to individuals~\cite{jain_50}. Since these traits are innate to an individual, they are regarded as more reliable than knowledge-based authentication schemes. On the downside, biometric authentication raises privacy concerns related to collecting  biometric data. In addition, spoof attacks  at  the biometric sensor level~\cite{phone_spoofing}, and possible theft of biometric templates stored inside the device, are among the growing concerns related to biometric-based authentication.

Although the use of explicit authentication schemes is widespread, they are both cumbersome and obtrusive as the user needs to actively focus on the authentication step before utilizing the device. For instance, an average user unlocks their phone around 80 times a day~\cite{phone_unlock}, which can be a source of frustration even for the most avid of users. It is also estimated that, on average, a smartphone user spends over 4 hours per day on their device~\cite{phone_per_day}. Unsurprisingly, more and more users prefer to set simple and weak passwords, increase the inactive period for lock-out time, or disable the authentication step completely~\cite{phone_disable1},~\cite{phone_disable2}. In addition, PIN codes, passwords, and biometric scans are well-suited for one-time authentication but are not effective in detecting intrusion after successful authentication by the genuine user when unlocking the phone. \emph{Passive} authentication systems tackle these challenges by providing an additional layer of security by frequently and unobtrusively monitoring the user's interaction with the device. In this paper, we propose a passive user authentication scheme for smartphones where users are not required to participate in any explicit authentication step. A brief summary of explicit and passive smartphone authentication methods is given in Table~\ref{tab:active_passive}.
%
%


\begin{table}[!t] 
\caption{Explicit vs. Passive Smartphone Authentication.}
{\renewcommand\arraystretch{1.25}}
\centering
\resizebox{\linewidth}{!}{
\begin{tabular}{|l|p{4cm}|p{4cm}|}
\noalign{\hrule height 1.5pt}
&\textbf{Explicit Authentication} & \textbf{Passive Authentication}\\ \hline
\textbf{Verification} & User explicitly provides credentials (password, PIN) or a biometric trait& User patterns (keystroke dynamics, touch gestures) are unobtrusively collected\\ \hline
\textbf{Enrollment} & Multiple samples of the biometric traits are required & `Template' generated from user patterns\\ \hline
\textbf{Advantages} & Fast and accurate & Seamless and non-intrusive\\
\noalign{\hrule height 1.5pt}
\end{tabular}
}
\label{tab:active_passive}
\end{table}

\begin{table*}[!t]
\footnotesize
\caption{A few related work on passive smartphone authentication.}
\centering
\begin{threeparttable}
\renewcommand{\arraystretch}{1.5}
\begin{tabularx}{\textwidth}{>{\centering\bfseries}l>{}X >{\centering}l >{\arraybackslash}X>{\arraybackslash}cc}
\noalign{\hrule height 1.5pt}
Study & Modality & Dataset Statistics & Classifier & Accuracy & Auth. Time\tnote{*}\\
   \noalign{\hrule height 1pt}
 HMOG~\cite{hmog} & Movement, tap, keys & 100 users, 24 sessions\tnote{$\dagger$} & Scaled Euclidean & EER 7.16\% & 60-120 seconds\\
   \hline
   Hold and Sign~\cite{burriro} & Movement, signature & 30 users\tnote{$\dagger$} & Multilayer Perceptron & 95\% TAR @ FAR = 3.1\% & 235 seconds\\
    \hline
    Touchalytics~\cite{touchalytics} & Touch gestures  & 41 users\tnote{$\dagger$} & kNN, SVM & EER 3.0\% & 11-43 seconds\\
    \hline
    Mahbub~\etal~\cite{umdaa} &  Movement and others & 48 users for 2 months & Hidden Markov Model & Accuracy 96.6\% & N/A\\
    \hline
    Fridman~\etal~\cite{fridman} & Stylometry, app \& web usage, GPS & 200 users for 5 months & SVM, n-gram & EER 5.0\% & 60 seconds\\
    \hline
    \rowcolor{gray!5}This study & 8 modalities in Table~\ref{tab:this_modalities} & 37 users for 15 days & Siamese LSTM & 97.15\% TAR @ FAR=0.1\% & 3 seconds\\
\noalign{\hrule height 1.5pt}
\end{tabularx}
\begin{tablenotes}\footnotesize
\item[] TAR = true accept rate; FAR = false accept rate; EER = equal error rate
\item[$\dagger$] No time span available for this study
\item[*] Time required before authentication
\end{tablenotes}
 \end{threeparttable}
\label{tab:related}
\end{table*}

The first and foremost difficulty in designing user authentication schemes for smartphones lies in gathering data from the wide array of sensors available in a smartphone, as well as from a variety of users. Smartphone users today are much more sensitive to their privacy and more aware of spywares, which may avert potential users from providing their data. In this paper, we acquired a dataset comprising of measurements from 30 sensor modalities for 37 users.

Besides data collection, a major challenge in designing a robust passive authentication system for smartphones involves extracting robust features from noisy\footnote{Smartphone sensors are prone to provide time-variant sources of noise leading to inaccurate measurements~\cite{noise}. } data. In addition, the robustness and accuracy of the authentication scheme needs to be thoroughly evaluated and inference should be performed in real-time. On average, a smartphone user's session lasts for 72 seconds~\cite{average_phone_use} and, therefore, the time required to authenticate the user should be as small as possible. It merely takes around 1.2 and 0.91 seconds to unlock an iPhone using FaceID and TouchID, respectively~\cite{iphone_speed}. 

We propose a Siamese Long Short-Term Memory (LSTM) architecture for extracting deep temporal features from the data corresponding to a number of passive sensors in smartphones for user authentication. We explore various operational modes and scenarios along with the corresponding authentication performance of the proposed method. 

Concisely, the contributions of the paper are as follows: 
\begin{itemize}
\itemsep0em 
    \item Proposed a passive user authentication method based on keystroke dynamics, GPS location, accelerometer, gyroscope, magnetometer, linear accelerometer, gravity, and rotation modalities that can unobtrusively verify a genuine user with $97.15\%$ TAR at 0.1\% FAR within 3 seconds.
    \item Acquired a dataset comprising of measurements from 30 different smartphone sensors for 37 users around the world. An Android application was designed to log data unobtrusively from the users' smartphones.
    \item Analyzed changes in accuracy when (1) multiple modalities are fused, and (2) authentication time is varied. Increasing the number of fused modalities boosts the accuracy, whereas the TAR at 0.1\% FAR drops from $99.80\%$ to $97.15\%$ for authentication times of 5 and 3 seconds, respectively.
\end{itemize}

\vspace{-0.8em}
\section{Related Work}
\subsection{Passive Smartphone Authentication}
Currently, there are around 2.5 billion active smartphone users in the world~\cite{number_phones}. With this increasing number, accurate, fast and robust authentication on smartphones has become an active area of research. Pioneering work on passive smartphone authentication was based on touchscreen analysis~\cite{feng},~\cite{lin}. Frank~\etal proposed a classification framework, namely Touchalytics, and achieved an EER of $4\%$ on a dataset comprising of 41 users using touchscreen input data. An obvious limitation of touchscreen recognition for passive authentication is the requirement of substantial explicit input from the user. 

Smartphones today are shipped with an array of sensors (see Figure~\ref{fig:sensors}). A topic of increasing number of studies has focused on passive smartphone authentication via motion sensors. In~\cite{learning_identity}, the authors proposed a continuous motion-based authentication system using data from accelerometer and gyroscope sensors. They obtained an EER of $18.2\%$, an error rate not suitable for security requirements. 

\subsection{Multimodal Biometric Systems}
Most of the passive authentication studies have focused on a single sensing modality for authentication. Authenticating a user on their smartphone based on a \emph{single} biometric modality becomes very challenging when the authentication time window is short. In addition, given the task the user is engaged in, the amount of data and the availability of different sensor modalities fluctuates. A robust passive authentication scheme must be able to adapt to the high intra-user variability observed in human-smartphone interaction. Sitova~\etal introduced a multimodal approach to passive smartphone authentication via accelerometer, gyroscope, and touch-screen observations~\cite{hmog}. Using a one-class SVM  classifier, an EER of $7.16\%$ was achieved; however, the authors deferred real-world scenarios, such as investigating authentication accuracy when the user is not engaged in typing, to future work. Niinuma~\etal explored a continuous authentication scheme using the user's face and color of clothing for verification using a webcam~\cite{niinuma}. In~\cite{crouse}, the authors explored a passive authentication system for smartphones using face recognition. However, unobtrusively acquiring face images may be invasive to the user's privacy. Using a Siamese convolution neural network, Centono~\etal showed a $97.8\%$ accuracy in verifying the genuine user~\cite{siamese} using accelerometer, gyroscope, and magnetometer sensor modalities. However, the study does not consider the temporal dependence between samples. DeepAuth, on the other hand, used a LSTM architecture to classify a genuine/impostor user via accelerometer, gyroscope, and magnetometer samples~\cite{amini}. However, they considered a small dataset where each user's session lasts for only 10-13 minutes. 

Fusing decisions from multiple modalities to authenticate the user has been demonstrated to be very useful~\cite{fusion}. The majority of multimodal biometric systems fuse classifiers at the score level based on min, max, or sum rules~\cite{combining}. In the proposed approach, we adopt the sum of scores fusion technique, which has been shown to perform well in multimodal biometric systems compared to other fusion schemes~\cite{simple_sum}. 

Limited studies on passive smartphone authentication have utilized multimodal biometric systems but, to the best of our knowledge, they have (1) all considered a small pool of modalities, (2) not evaluated the temporal performance of intrusion detection, and (3) not considered the temporal dependence of features across modalities. In this study, we propose a Siamese LSTM network to address temporal dependencies. A brief list of related work on passive smartphone authentication is given in Table~\ref{tab:related}.
\section{System Overview}
In the off-line phase, an authentication model is trained via the proposed methodology for each modality. During deployment, the incoming data from the smartphone sensor modalities are continuously monitored. If the incoming data successfully passes the authentication criteria, a decision is made that the current user is indeed the legitimate owner of the devices. Otherwise, the system locks out the user from the device and offers an explicit authentication method such as a password, or biometrics such as fingerprint scans.
%

\section{Dataset}
The dataset used in this work consists of measurements from 30 sensors (see Table~\ref{tab:all_modalities} for all 30 modalities), currently present in most commonly used smartphones (see Table~\ref{tab:phones}), for 37 users. Data for each user was collected over a period of 15 days. Users that participated in the data collection process were primarily students from universities, across different countries, who are also regular smartphone users\footnote{The users were contacted by the authors and a consent form along with the link to the data collection application was sent.}. Dataset statistics are given in Table~\ref{tab:dataset}. 

\begin{table}[!t] 
\caption{The eight dominant sensor modalities considered in our study. All the 30 modalities collected can be found in Table~\ref{tab:all_modalities}.}
\centering
\footnotesize
\resizebox{\linewidth}{!}{
\begin{tabular}{|l|l|}
\noalign{\hrule height 1.5pt}
Keystroke Dynamics    & Key hold time, finger area and finger pressure\\ \hline
GPS Location & User's GPS location (latitude, longitude)\\ \hline
Accelerometer & Smartphone's acceleration in X, Y, Z plane\\ \hline
Gyroscope Gesture & Rate of rotation of the device in X, Y, and Z planes\\ \hline
Magnetometer   & Earth's magnetic field in X, Y, and Z planes\\ \hline
Gravity Sensor & Direction and magnitude of gravity\\ \hline
Linear Acceleration & Linear acceleration in X, Y, and Z planes\\ \hline
Rotation Sensor & Device's rotation in X, Y, and Z planes\\ \hline
\noalign{\hrule height 1.5pt}
\end{tabular}
}
\label{tab:this_modalities}
\end{table}

\begin{table}[!t] 
\footnotesize
\caption{Dataset Statistics. A record is a measurement for a sensor modality. Users that participated in the study are primarily students located in different regions of the world, including USA, India, Turkey, Brazil, and Dominican Republic.}
\centering
\begin{tabular}{|l|c|}
\noalign{\hrule height 1.5pt}
 No. of users & 37 \\ \hline
 Data collection duration & 15 days\\ \hline
 No. of sensor modalities & 30\\ \hline
 Total number of records &  6.7M\\ \hline
 Average number of records per user & 180K\\ \hline
 Male to Female Ratio (\%) & 57/43\\ \hline
 Age Range (years) & 18 - 56\\
  \hline
\noalign{\hrule height 1.5pt}
\end{tabular}
\label{tab:dataset}
\end{table}

An Android application\footnote{[URL ommited for double-blind review]} was built that passively acquired data from the sensors (see Figure~\ref{fig:app}). This application automatically turns on whenever the smartphone boots up and continuously runs in the background while passively recording sensor data. In order to collect keystroke dynamics, we also built a custom soft-keyboard, installable from the data collection application. 

To the best our knowledge, our dataset is unique due to (i) its rich sensor space (30 different sensors), and (ii) the manner in which data was acquired keeping the real-world scenario in mind. First, no user interaction with the data collection application is required, thereby enabling the users to use their smartphones as they generally would in their everyday lives. In order to simulate real world performance, data from users were collected from their own personal devices with no restrictions placed on the usage patterns, the Android device, or the Android OS version. Data from the modalities were acquired continuously, even when the user is not actively interacting with their smartphone.

In this study, we evaluate authentication performance of our proposed method on eight modalities (see Table~\ref{tab:this_modalities}), (i) keystroke dynamics, (ii) GPS location, and (iii) accelerometer, (iv) gyroscope, (v) magnetometer, (vi) linear acceleration, (vii) gravity, and (viii) rotation. We chose these modalities due to their popularity in literature~\cite{antal_2},~\cite{hmog},~\cite{fridman}, and because these sensors are available in all smartphones. In addition, among all the 30 modalities, measurements from these sensors are most abundant and distinctive.

\paragraph{Keystroke Dynamics} By modeling user's typing rhythms and mannerisms, can be used for authenticating smartphone users. We record the finger pressure, finger area, and hold time whenever a user types a character on their smartphone (similar to~\cite{antal_2}). The exact characters typed are not logged and therefore, the keystroke patterns collected are non-invasive in nature.

\paragraph{GPS Location} For every user in the dataset, a pair of latitude and longitude coordinates is recorded whenever the device was moved. Location is considered as a measure of an individual's characteristic and, therefore, we hypothesize that distinguishable patterns can be found in a user's everyday location.

\begin{figure*}[!t]
\centering
\includegraphics[width=0.75\linewidth]{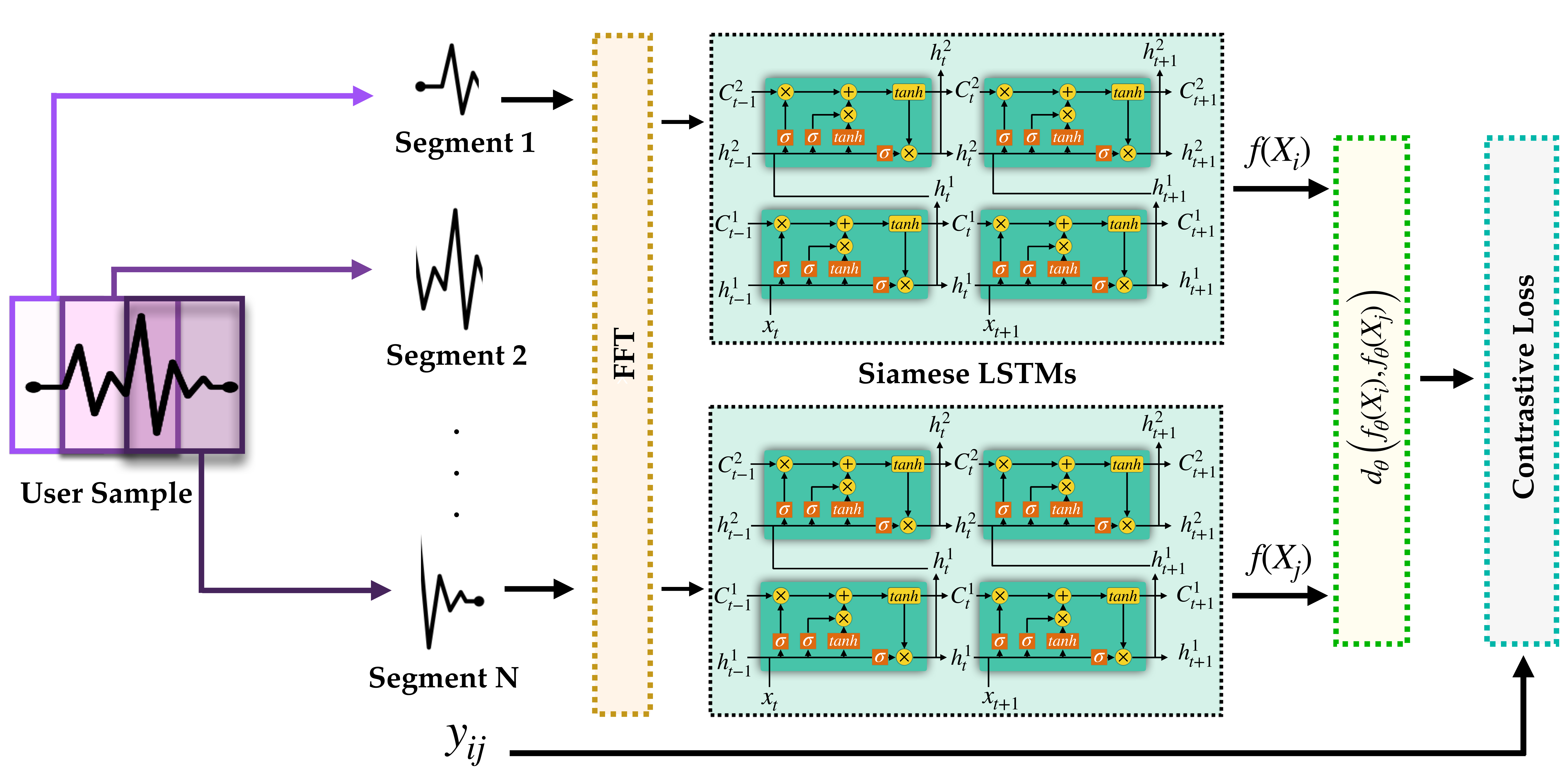}
\caption{Architecture of the proposed model. Here, $\{X_i, X_j\}$ are input segment pairs, $y_{ij}$ is the label, and $f_\theta(\cdot)$ and $d_\theta$ denote the embedding and distance function, respectively. The parameters of the Siamese LSTM network are denoted by $\theta$.}
\label{fig:architecture}
\end{figure*}

\paragraph{Movement} We evaluate the authentication performance for all the six movement sensors in our dataset: accelerometer, gyroscope, magnetometer, linear accelerometer, gravity, and rotation. Measurements are recording in three axes, $X$, $Y$, and $Z$, for all six sensors. The sampling rate for all six sensor data stream is 1 Hz,~\ie one measurement per second. 


In our dataset, intra-user chronological gaps in measurements exists to the high-variability in user's behavior, for instance, they may switch off their smartphone, or the phone may shut off due to battery drain. Even though these intra-user variations in the dataset pose additional challenges, it also better simulates real-world scenarios where such discrepancies cannot be avoided. 

\section{Methodology}
As shown in Figure~\ref{fig:sensors}, smartphones today are shipped with an array of sensors including global positioning system (GPS), accelerometer, gyroscope, magnetometer, and others. In this paper, our goal is to utilize data from these modalities in order to verify whether the genuine owner of the device is logged in. There are two main passive smartphone authentication model training strategies, namely,~\emph{on-line} and~\emph{off-line}. On-line approach trains an authentication model using samples pertaining to the smartphone user for a certain period of time before deploying the model. A major limitation to on-line approach involves training an individual model for each user. As a consequence, it is challenging to accurately evaluate the overall authentication performance across all the users due to high variance. In addition, the required amount of data, the duration of data collection before model deployment, and privacy concerns of storing the training data are ongoing challenges. 

Off-line approaches, on the other hand, train a common authentication model that learns salient representations for individual modalities. In this approach, the same trained model is deployed when users install the application. Moreover, the users can avail of the authentication mechanism immediately after the installation of the application. For these reasons, we propose an off-line learning strategy for passive smartphone authentication. In particular, for each of the eight modalities, we train a Siamese LSTM network to learn deep temporal features. Samples from users are transformed into an embedding space learned by the Siamese network. Therefore, during deployment, only the features extracted from the incoming data are required for authentication, thereby, eliminating the need to store private data on the device. The proposed method is comprised of three modules: (1) data sampling, (2) preprocessing, and (3) Siamese LSTM. Figure~\ref{fig:architecture} outlines the overall architecture of the proposed method.

\paragraph{Sampling Module} Suppose that we extract a $D$-dimensional data sample for a given sensor modality (for instance, accelerometer has data in 3 axes, namely $X$, $Y$, and $Z$). The number of samples for each user can vary, and our dataset is comprised of chronologically irregular measurements for the same user due to various reasons such as their phones being switched off. Therefore, we segment the data by moving a window of fixed size $T$ (authentication time window) over the sequential data with a pre-defined shift of $T_{shift}$ and build overlapping fixed-sized segments. Hence, for each user, we have a set of $D\times T$ segments. These segments are then passed to the \emph{preprocessing} module.

\paragraph{Preprocessing Module}
The outputs of the sampling module contain measurements from a modality in their original domain, namely the time domain. The frequency domain can handle and remove noise, while also retaining the discriminating patterns in the data within sequential data~\cite{frequency_domain}. We map measurements from the time domain to frequency domain only for the movement sensors,~\ie accelerometer, gyroscope, magnetometer, linear accelerometer, gravity, and rotation. Fast Fourier Transform (FFT)~\cite{fft} is utilized to convert time domain signals on each feature dimension to frequency domain signals. The output of the FFT vectors are concatenated with samples in the time domain so that we can utilize information from both the domains.

\begin{table*}[!t] 
\footnotesize
\caption{Authentication performance of the 8 modalities. Across 5 folds, the mean and std. dev. of the TARs at 1.0\% and 0.1\% FAR are given. The best performing method is highlighted in light gray. }
\centering
\resizebox{\linewidth}{!}{
\begin{tabular}{l||a|a||c|c||c|c||c|c}
\noalign{\hrule height 1.5pt}
\textbf{Modality} & \multicolumn{2}{c||}{\textbf{Proposed Siamese LSTM}} & \multicolumn{2}{c||}{\textbf{Siamese CNN~\cite{siamese}}} & \multicolumn{2}{c||}{\textbf{LSTM~\cite{amini}}} & \multicolumn{2}{c}{\textbf{Euclidean Distance}}\\ \hline
& \textbf{1.0\% FAR} & \textbf{0.1\% FAR} & \textbf{1.0\% FAR} & \textbf{0.1\% FAR} & \textbf{1.0\% FAR} & \textbf{0.1\% FAR}& \textbf{1.0\% FAR} & \textbf{0.1\% FAR}\\
 \hline
 Keystroke Dynamics & 81.61 $\pm$ 13.65 & 58.71 $\pm$ 14.61 & 71.12 $\pm$ 15.67 & 43.87 $\pm$ 11.39 & 59.87 $\pm$ 18.82 & 26.73 $\pm$ 16.75 & 12.11 $\pm$ 7.60 & 8.20 $\pm$ 7.60\\
 GPS & 78.34 $\pm$ 4.76 & 52.21 $\pm$ 5.76 & 63.23 $\pm$ 10.82 & 39.23 $\pm$ 8.26 & 51.87 $\pm$ 9.97 & 21.42 $\pm$ 7.76 & 21.32 $\pm$ 1.32 & 12.76 $\pm$ 1.65\\
 Accelerometer & 74.56 $\pm$ 5.64 & 37.74 $\pm$ 6.67 & 67.28 $\pm$ 6.61 & 35.98 $\pm$ 7.33 & 64.83 $\pm$ 3.73 & 23.11 $\pm$ 3.82 & 13.06 $\pm$ 1.87 & 8.01 $\pm$ 0.81\\
 Gyroscope & 44.15 $\pm$ 7.53 & 15.18 $\pm$ 3.50 & 28.14 $\pm$ 7.69 & 11.33 $\pm$ 2.21 & 36.68 $\pm$ 7.43 & 8.89 $\pm$ 2.39 & 8.15 $\pm$ 1.23 & 6.54 $\pm$ 0.81\\
 Magnetometer  & 74.15 $\pm$ 7.52 & 46.19 $\pm$ 8.83 & 60.91 $\pm$ 9.07 & 32.21 $\pm$ 6.87 & 26.33 $\pm$ 9.26 & 10.26 $\pm$ 8.33 & 18.85 $\pm$ 5.15 & 16.51 $\pm$ 5.88\\
  Linear Accelerometer  & 50.19 $\pm$ 14.86& 28.39 $\pm$ 16.20 & 46.35 $\pm$ 17.62 & 27.97 $\pm$ 18.74 & 29.89 $\pm$ 9.54 & 11.53 $\pm$ 6.78  & 8.91 $\pm$ 1.38 & 7.66 $\pm$ 0.95 \\
   Gravity  & 69.95 $\pm$  4.35 & 32.24 $\pm$ 2.49 & 61.87 $\pm$ 7.95 & 31.92 $\pm$ 4.32 & 52.43 $\pm$ 5.64 & 32.12 $\pm$ 4.14 & 18.07 $\pm$ 5.41 & 10.98 $\pm$ 3.36\\
    Rotation  & 74.85 $\pm$ 4.78 & 41.52 $\pm$ 3.02 & 61.91 $\pm$ 4.14 & 30.08 $\pm$ 1.29 & 56.75 $\pm$ 4.86 & 35.21 $\pm$ 3.98 & 17.96 $\pm$ 5.6 & 13.33 $\pm$ 3.72\\
  \hline
\noalign{\hrule height 1.5pt}
\end{tabular}
}
\label{tab:ind}
\end{table*}

\paragraph{Siamese LSTM} 
Our goal is to obtain highly discriminative features for each modality that can distinguish samples from genuine and impostor users. In other words, we would like to learn information-rich transformation of the data from modalities into an embedding space that can preserve distance relation between training samples. Suppose we are given a pair of input samples,  $\{X_i, X_j\}$. Let $y_{ij}$ be a label, such that, $y_{ij} = 0$, if $X_i$ and $X_j$ belong to the same user, and $y_{ij}=1$, otherwise. Our objective is to map input samples to an embedding space where two samples from the same user are closer together and two samples from different users are far apart. A Siamese network architecture, which is a neural network architecture comprising of two identical sub-networks, is well-suited for such verification tasks~\cite{siamese1},~\cite{contrastive}. In this manner, relationship between two input samples can be learned. In a Siamese network, weights between the two sub-networks are shared and the weights are updated based on the label, $y_{ij}$. 

A Siamese Convolutional Neural Network (CNN) was previously proposed for passive smartphone authentication~\cite{siamese}; however, CNNs are not well-suited to capture the temporal dependence within samples. We leverage Long Short-Term Memory (LSTM)~\cite{lstm} as our Siamese architecture to model patterns in users' data. LSTM, a variant of Recurrent Neural Networks (RNN), is designed for classifying, processing and making predictions on time series data. In our approach, we stack two LSTMs in order to learn hierarchical representation of the time series data. The first LSTM outputs a sequence of vectors, ${h^1_{1},\ldots,h^1_{T}}$ which are then fed as input to the second LSTM. The last hidden state, $h^2_{T}$, of the second LSTM represents the final non-linear embedding, denoted by $f_{\theta}(\cdot)$, where $\theta$ represents the parameters of the Siamese LSTM network. This hierarchy of hidden layers allows for more salient representation of the time-series data. We denote the embedding size of both LSTMs as $C$.

In order to train the Siamese LSTM network, we define a pairwise contrastive loss function. For a given pair of input samples, the Euclidean distance between the two output feature vectors from the two sub-networks are fed to the contrastive loss function. This loss function regulates large or small distances depending on the label associated with the pair of samples, $y_{ij}$. In this manner, we ensure that the Euclidean distance between the pairs, $d_{\theta}(X_i, X_j)$, where,
\begin{align*}
d_{\theta}(X_i, X_j)  = ||f_{\theta} (X_i) - f_{\theta} (X_j)||_2,
\end{align*}
is small for genuine pairs and large for impostor pairs. The contrastive loss function is defined as~\cite{contrastive}:
\begin{align*}
\ell_{\theta} &= \Sigma_{i,j = 1}^{N}L_{\theta}(X_i, X_j, y_{ij}),~\quad\quad\text{where}\\
L_{\theta} &= (1 - y_{ij})\frac{1}{2}(d_{\theta})^2 + (y_{ij})\frac{1}{2}\left\{max(0, \alpha - d_{\theta})\right\}^2
\end{align*}
where, $\alpha > 0$ is called the margin.

\section{Experimental Results}
The contrastive loss function is optimized using Adam~\cite{adam} optimizer. The embedding size is fixed at 16. Segment shift is set to $T_{shift} = 1$ for all the experiments. Training details can be found in Appendix~\ref{appndx:details}.

\subsection{Individual Modality Performance}
\label{sec:ind}
Since a separate Siamese LSTM model is trained for each modality, we first evaluate the authentication performance of each individual modality. We perform 5-fold cross-validation such that each fold comprises of 29 users for training and 8 users for testing. The proposed method is trained and tested on 20-second segments ($T = 20$). In order to train and evaluate our model, pairs of segment samples are generated. Genuine pairs are all possible pairs of segments from the same user. A sample from a user is paired with another user's segment randomly to constitute an impostor pair. On average, each user has around 180,000 genuine  pairs and the same number of impostor pairs across all the modalities.

In Table~\ref{tab:ind}, authentication performance of the proposed method is compared with three baselines, (1) Siamese CNN proposed by Centeno~\etal~\cite{siamese}, (2) a single LSTM network proposed by Amini~\etal~\cite{amini}, and (3) Euclidean distance classifier. The proposed Siamese LSTM network outperforms Siamese CNN due to LSTM's capability of capturing temporal dependencies. In addition, Siamese architectures are better suited for preserving distances between pairs of input samples. Thus, the proposed architecture has a better authentication performance than a single LSTM. Since  Euclidean distance is used in the proposed method, we also investigate the authentication performance without learning a non-linear transformation of the temporal data. For this purpose, we obtain thresholds at $1.0\%$ and $0.1\%$ FARs by computing the Euclidean distances between the raw segment pairs in the training set. These thresholds are used to compute the TARs of the Euclidean distances between the raw testing segment pairs. We observe that using deep temporal features significantly improves the authentication performance.

\begin{figure}
\centering
\includegraphics[width=0.65\linewidth]{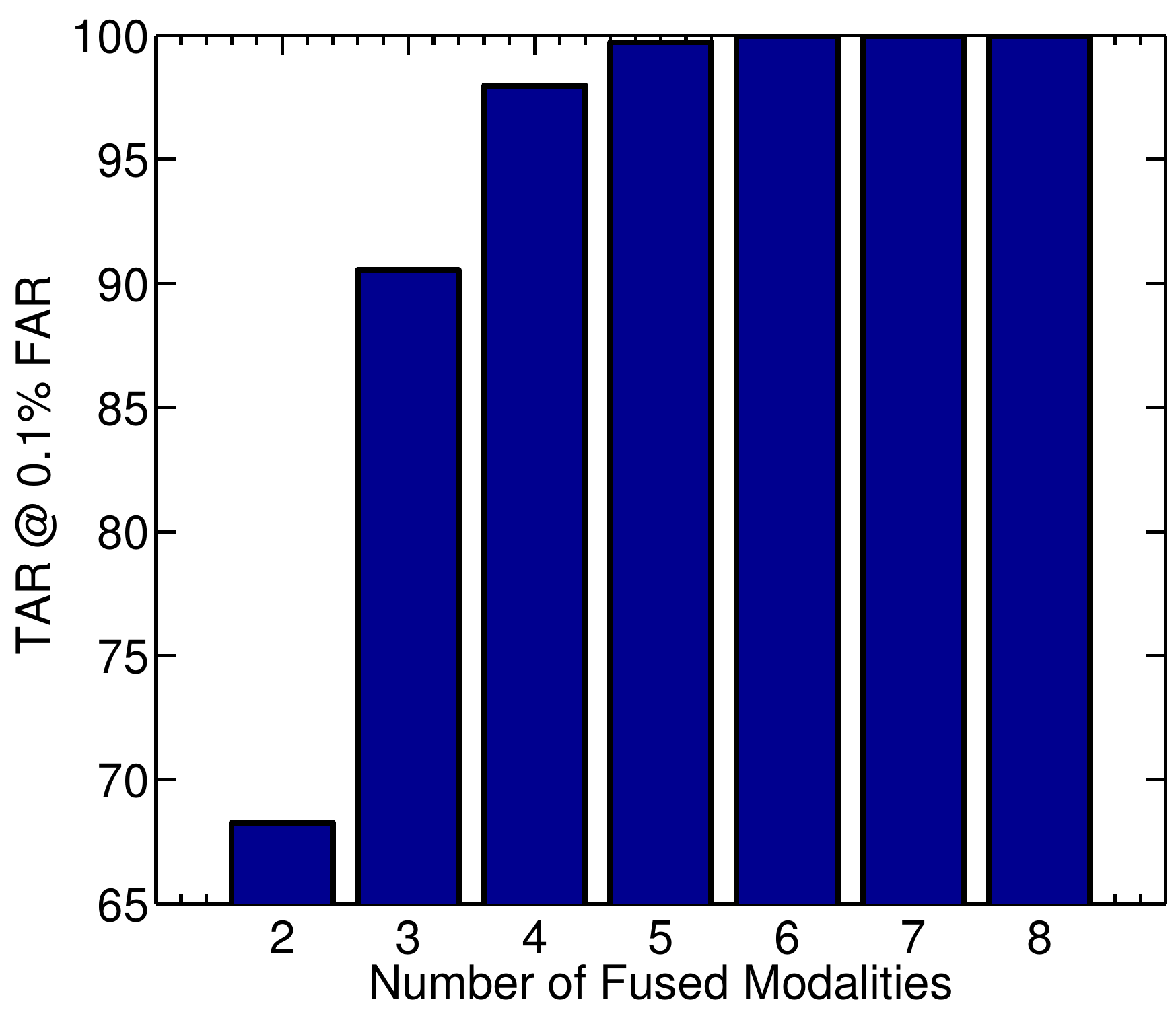}
\caption{Average TAR at 0.1\% FAR on fusing all possible subsets of modalities. Increasing the number of fused modalities boosts the overall accuracy up to 5 modalities.}
\label{fig:fusion}
\end{figure}

\begin{table}[!t]
\centering
\resizebox{\linewidth}{!}{
\footnotesize
\begin{tabular}{|c|l|c|}
\noalign{\hrule height 1.0pt}
& \textbf{Modalities} & \textbf{TAR (\%) @ 0.1\% FAR}\\
\noalign{\hrule height 0.8pt}
\parbox[p]{2mm}{\multirow{3}{*}{\rotatebox[origin=c]{90}{\textbf{Top-3}}}} & Acc + Grv + Gyr + Key + Lin + Mag  + GPS & 99.92  $\pm$ 0.07 \\
& Acc + Grv + Gyr + Lin + Mag + Rot + GPS & 99.96 $\pm$ 0.02 \\
& All & 99.98 $\pm$ 0.01\\
\noalign{\hrule height 0.5pt}
\parbox[t]{2mm}{\multirow{3}{*}{\rotatebox[origin=c]{90}{\textbf{Worst-3}}}} & Gyr + Key & 50.94 $\pm$ 29.05\\
& Key + Lin & 55.11 $\pm$ 31.43\\
& Gyr + Lin & 48.25 $\pm$ 4.13\\ 
\noalign{\hrule height 1.0pt}
\end{tabular}
}
\caption{Top-3 best and worst-3 performing subsets of sensing modalities across 5 folds\protect\footnotemark.}
\label{tab:fusion}
\end{table}

\footnotetext{Here, Key,  GPS, Acc,  Gyr, Mag, Lin, Grv, Rot, and All refer to keystroke dynamics, GPS location, accelerometer, gyroscope, magnetometer, linear accelerometer, gravity, rotation, and all eight modalities combined, respectively.}

\subsection{Fusion of Modalities}
\label{sec:fusion}
The performance of the individual modalities is far from satisfactory. Given a short authentication time window, relying on a single modality for authentication is, therefore, not practical for robust and accurate authentication. In addition to poor performance of individual modalities, all the modalities may not even be available at any given time of the day. For instance, if the user is not using the keyboard, we will not have access to keystroke dynamics. Therefore, for a reliable passive authentication system, it is imperative that we rely on multiple modalities. We evaluate the authentication performance for all possible subsets of modalities on fusing 2, 3, 4, 5, 6, 7, and 8 modalities together (in total, there are 247 different possible subsets). The corresponding genuine and impostor scores for the modality subsets are fused using the sum score fusion technique.
%

The average TARs at 0.1\% FAR for all the possible subsets are shown in Figure~\ref{fig:fusion}. Table~\ref{tab:fusion} gives the top-3 best and worst-3 performing modality subsets. It is observed that:
(1) increasing the number of fused modalities boosts the overall authentication performance; (2) the highest accuracy is achieved (99.98\% TAR @ 0.1\% FAR) when all eight modalities are fused together; (3) the performance saturates after fusing 5 different modalities.

\subsection{Modality Contribution}
\label{sec:contrib}

\begin{figure}[!t]
\centering
\includegraphics[width=0.65\linewidth]{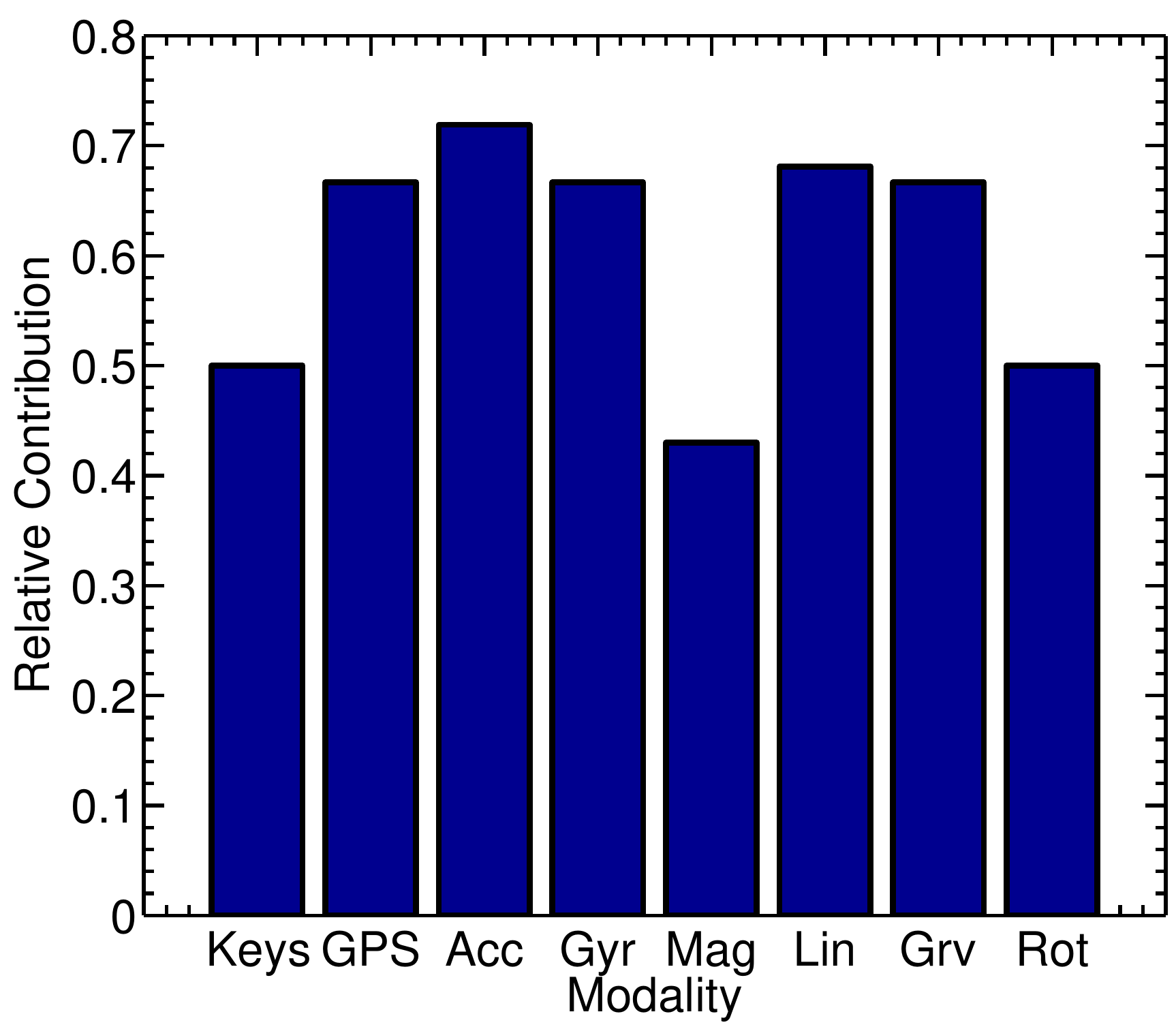}
\caption{Relative contribution of each modality.}
\label{fig:contribution}
\end{figure}

The next natural question to ask is, \emph{`Which modality contributes the most when all the 8 modalities are fused?'}. Let $M = {m_1,\ldots,m_8}$ be the set of all modalities. When a modality, $m_{i}$, is not considered in the fusion, the drop in overall authentication performance is computed by $\left(TAR_{M} - TAR_{M'}\right)$, where $M'$ contains all the modalities except $m_{i}$ and $TAR_{M}$ denotes the True Accept Rate (\%) at 0.1\% FAR on fusing all modalities in $M$. The contribution for modality, $m_i$, is defined as $\left(TAR_{M} - TAR_{M'}\right)/(100-TAR_{M'})$. We plot the relative contributions of the eight modalities in Figure~\ref{fig:contribution}. Note that a high performing modality may not necessarily have a high contribution. For instance, magnetometer has the lowest contribution to the overall authentication performance out of the eight modalities. Linear accelerometer, on the other hand, has a relatively high contribution even though it performs poorly on its own. This is likely because the comparison scores obtained from the linear accelerometer model is complementary to scores from other modalities. 

\subsection{Temporal Information}
\label{sec:temp}

\begin{figure}[!t]
\centering
\includegraphics[width=0.65\linewidth]{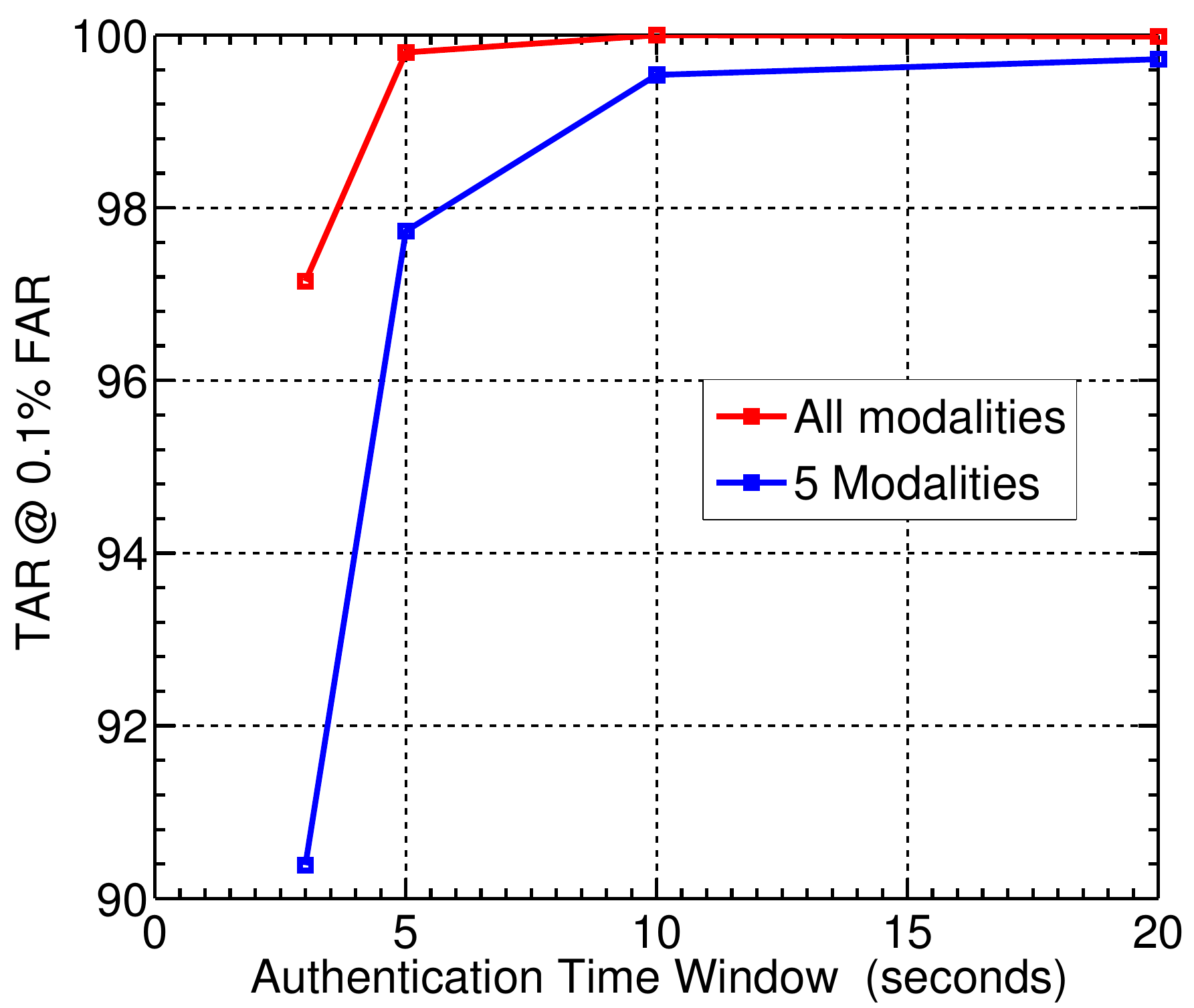}
\caption{TAR at $0.1\%$ FAR for authentication time windows varying from 3 to 20 seconds.}
\label{fig:temporal}
\end{figure}

A common phenomenon in passive authentication systems involves a trade-off between the authentication time and recognition performance. Figure~\ref{fig:temporal} shows the authentication performance for segment sizes of $T = 3$, $5$, $10$, and $20$ seconds. Indeed, we find that accuracy drops with decreasing authentication time windows, likely due to lack of information content required to successfully authenticate the user. Therefore, the window size can be chosen depending on the application at hand. When shorter authentication times are desired, then more number of modalities should be considered for a secure application.

\section{Conclusions}
We have proposed a Siamese LSTM architecture for passive authentication of smartphone users. We collected a dataset comprised of measurements from 30 sensor modalities, for 37 smartphone users, over a time period of 15 days. We evaluated the authentication performance under various scenarios for 8 dominant modalities, namely keystroke dynamics, GPS location, accelerometer, gyroscope, magnetometer, linear accelerometer, gravity, and rotation. We summarize our findings as follows:
\begin{itemize}
\itemsep0em 
    \item The proposed method can passively authenticate a smartphone user with TAR of $99.80\%$ and $97.15\%$ at 0.1\% FAR within 5 and 3 seconds, respectively. 
    \item Relying on a single modality for authentication is not reliable. Fusing a subset of modalities boosts the authentication accuracy. While the highest authentication accuracy is achieved when all eight modalities are fused together (99.98\% TAR @ 0.1\% FAR), a subset of 5 modalities results in comparable recognition performance (99.71\% TAR @ 0.1\% FAR). 
\end{itemize}

With the growing number of sensors found in smartphones, it is important to explore robust and unobtrusive passive authentication approaches. In the future, we plan on fusing additional smartphone modalities. In addition, we will also design a continuous authentication scenario where our methodology can be deployed.

\section*{Acknowledgement}
This research was supported by Ford Motor Company. The authors would like to thank Dr. Kwaku O. Prakah-Asante and Dr. K. Venkatesh Prasad from Ford who provided insight and expertise that greatly assisted the research.

{\small
\bibliographystyle{ieee}
\footnotesize{
\bibliography{egbib}}

\begin{thebibliography}{10}\itemsep=-1pt

\bibitem{iphone_speed}
{9To5Mac}.
\newblock {iPhone X Face ID versus Touch ID — which is faster?}
\newblock \url{https://9to5mac.com/2017/11/01/face-id-versus-touch-id-speed},
  2017.

\bibitem{amini}
S.~Amini, V.~Noroozi, A.~Pande, S.~Gupte, P.~S. Yu, and C.~Kanich.
\newblock Deepauth: A framework for continuous user re-authentication in mobile
  apps.
\newblock In {\em Proceedings of the 27th ACM International Conference on
  Information and Knowledge Management}, pages 2027--2035. ACM, 2018.

\bibitem{antal_2}
M.~Antal and L.~Z. Szab{\'o}.
\newblock An evaluation of one-class and two-class classification algorithms
  for keystroke dynamics authentication on mobile devices.
\newblock In {\em Control Systems and Computer Science (CSCS), 2015 20th
  International Conference on}, pages 343--350. IEEE, 2015.

\bibitem{delta_ID}
{Biometric Update}.
\newblock {Fujitsu’s latest smartphone using Delta ID iris recognition
  technology}.
\newblock \url{https://bit.ly/2CMzDII}, 2016.

\bibitem{siamese1}
J.~Bromley, I.~Guyon, Y.~LeCun, E.~S{\"a}ckinger, and R.~Shah.
\newblock Signature verification using a" siamese" time delay neural network.
\newblock In {\em Advances in neural information processing systems}, pages
  737--744, 1994.

\bibitem{burriro}
A.~Buriro, B.~Crispo, F.~Delfrari, and K.~Wrona.
\newblock Hold and sign: a novel behavioral biometrics for smartphone user
  authentication.
\newblock In {\em Security and Privacy Workshops (SPW), 2016 IEEE}, pages
  276--285. IEEE, 2016.

\bibitem{phone_unlock}
{Business Insider}.
\newblock {The average iPhone is unlocked 80 times per day}.
\newblock \url{https://read.bi/2EcOzkb}, 2016.

\bibitem{siamese}
M.~P. Centeno, Y.~Guan, and A.~van Moorsel.
\newblock {Mobile Based Continuous Authentication Using Deep Features}.
\newblock 2018.

\bibitem{fft}
J.~W. Cooley and J.~W. Tukey.
\newblock An algorithm for the machine calculation of complex fourier series.
\newblock {\em Mathematics of Computation}, 19(90):297--301, 1965.

\bibitem{crouse}
D.~Crouse, H.~Han, D.~Chandra, B.~Barbello, and A.~K. Jain.
\newblock Continuous authentication of mobile user: Fusion of face image and
  inertial measurement unit data.
\newblock In {\em Biometrics (ICB)}, pages 135--142. IEEE, 2015.

\bibitem{phone_guess}
{Data Genetics}.
\newblock {PIN analysis}.
\newblock \url{http://www.datagenetics.com/blog/september32012/}, 2012.

\bibitem{feng}
T.~Feng, Z.~Liu, K.-A. Kwon, W.~Shi, B.~Carbunar, Y.~Jiang, and N.~Nguyen.
\newblock Continuous mobile authentication using touchscreen gestures.
\newblock In {\em Homeland Security (HST), 2012 IEEE Conference on Technologies
  for}, pages 451--456. Citeseer, 2012.

\bibitem{touchalytics}
M.~Frank, R.~Biedert, E.~Ma, I.~Martinovic, and D.~Song.
\newblock Touchalytics: On the applicability of touchscreen input as a
  behavioral biometric for continuous authentication.
\newblock {\em IEEE Transactions on Information Forensics and Security},
  8(1):136--148, 2013.

\bibitem{fridman}
L.~Fridman, S.~Weber, R.~Greenstadt, and M.~Kam.
\newblock Active authentication on mobile devices via stylometry, application
  usage, web browsing, and gps location.
\newblock {\em IEEE Systems Journal}, 11(2):513--521, 2017.

\bibitem{phone_per_day}
{Hackernoon}.
\newblock {How Much Time Do People Spend on Their Mobile Phones in 2017?}
\newblock
  \url{https://hackernoon.com/how-much-time-do-people-spend-on-their-mobile-phones-in-2017-e5f90a0b10a6},
  2017.

\bibitem{noise}
I.~Hazan and A.~Shabtai.
\newblock Noise reduction of mobile sensors data in the prediction of
  demographic attributes.
\newblock In {\em Proceedings of the Second ACM International Conference on
  Mobile Software Engineering and Systems}, pages 117--120. IEEE Press, 2015.

\bibitem{lstm}
S.~Hochreiter and J.~Schmidhuber.
\newblock Long short-term memory.
\newblock {\em Neural Computation}, 9(8):1735--1780, 1997.

\bibitem{simple_sum}
A.~Jain, K.~Nandakumar, and A.~Ross.
\newblock Score normalization in multimodal biometric systems.
\newblock {\em Pattern Recognition}, 38(12):2270--2285, 2005.

\bibitem{jain_50}
A.~K. Jain, K.~Nandakumar, and A.~Ross.
\newblock 50 years of biometric research: Accomplishments, challenges, and
  opportunities.
\newblock {\em Pattern Recognition Letters}, 79:80--105, 2016.

\bibitem{adam}
D.~P. Kingma and J.~Ba.
\newblock Adam: A method for stochastic optimization.
\newblock {\em arXiv preprint arXiv:1412.6980}, 2014.

\bibitem{combining}
J.~Kittler, M.~Hatef, R.~P. Duin, and J.~Matas.
\newblock On combining classifiers.
\newblock {\em IEEE Transactions on Pattern Analysis and Machine Intelligence},
  20(3):226--239, 1998.

\bibitem{lin}
C.-C. Lin, C.-C. Chang, D.~Liang, and C.-H. Yang.
\newblock A new non-intrusive authentication method based on the orientation
  sensor for smartphone users.
\newblock In {\em Software Security and Reliability (SERE), 2012 IEEE Sixth
  International Conference on}, pages 245--252. IEEE, 2012.

\bibitem{umdaa}
U.~Mahbub, S.~Sarkar, V.~M. Patel, and R.~Chellappa.
\newblock Active user authentication for smartphones: A challenge data set and
  benchmark results.
\newblock In {\em Biometrics Theory, Applications and Systems (BTAS)}, pages
  1--8. IEEE, 2016.

\bibitem{learning_identity}
N.~Neverova, C.~Wolf, G.~Lacey, L.~Fridman, D.~Chandra, B.~Barbello, and
  G.~Taylor.
\newblock Learning human identity from motion patterns.
\newblock {\em IEEE Access}, 4:1810--1820, 2016.

\bibitem{average_phone_use}
{Nielsen Norman Group}.
\newblock {Mobile User Experience: Limitations and Strengths}.
\newblock \url{https://www.nngroup.com/articles/mobile-ux/}.

\bibitem{niinuma}
K.~Niinuma and A.~K. Jain.
\newblock Continuous user authentication using temporal information.
\newblock In {\em Biometric Technology for Human Identification VII}, volume
  7667, page 76670L. International Society for Optics and Photonics, 2010.

\bibitem{phone_disable1}
{Norton}.
\newblock {Norton Survey Reveals One in Three Experience Cell Phone Loss,
  Theft}.
\newblock
  \url{https://www.symantec.com/about/newsroom/press-releases/2011/symantec_0208_01},
  2011.

\bibitem{smartphone_most_common}
{Ofcom}.
\newblock {Communications Market Reports}.
\newblock
  \url{https://www.ofcom.org.uk/research-and-data/multi-sector-research/cmr},
  2018.

\bibitem{frequency_domain}
L.~R. Rabiner and B.~Gold.
\newblock {Theory and Application of Digital Signal Processing}.
\newblock {\em Englewood Cliffs, NJ, Prentice-Hall, Inc.}, 1975.

\bibitem{steal_info}
{Scott Wright}.
\newblock {The Symantec Smartphone Honey Stick Project}.
\newblock \url{https://symc.ly/2AsSDdz}, 2012.

\bibitem{phone_spoofing}
{Security Research Labs}.
\newblock {Fingerprints are not fit for secure device unlocking}.
\newblock \url{https://srlabs.de/bites/spoofing-fingerprints/}.
\newblock [Online; accessed 24-November-2018].

\bibitem{fusion}
T.~Sim, S.~Zhang, R.~Janakiraman, and S.~Kumar.
\newblock Continuous verification using multimodal biometrics.
\newblock {\em IEEE Transactions on Pattern Analysis and Machine Intelligence},
  29(4):687--700, 2007.

\bibitem{hmog}
Z.~Sitov{\'a}, J.~{\v{S}}ed{\v{e}}nka, Q.~Yang, G.~Peng, G.~Zhou, P.~Gasti, and
  K.~S. Balagani.
\newblock Hmog: New behavioral biometric features for continuous authentication
  of smartphone users.
\newblock {\em IEEE Transactions on Information Forensics and Security},
  11(5):877--892, 2016.

\bibitem{phone_disable2}
{Sophos}.
\newblock {Survey says 70\% don't password-protect mobiles}.
\newblock
  \url{https://nakedsecurity.sophos.com/2011/08/09/free-sophos-mobile-security-toolkit/},
  2016.

\bibitem{number_phones}
{Statista}.
\newblock {Number of smartphone users worldwide from 2014 to 2020 (in
  billions)}.
\newblock \url{https://read.bi/2EcOzkb}, 2018.

\bibitem{contrastive}
Y.~Taigman, M.~Yang, M.~Ranzato, and L.~Wolf.
\newblock Closing the gap to human-level performance in face verification.
  deepface.
\newblock In {\em IEEE Computer Vision and Pattern Recognition (CVPR)}, 2014.

\bibitem{def_smartphone}
{Wikipedia}.
\newblock {Smartphone}.
\newblock \url{https://en.wikipedia.org/wiki/Smartphone}, 2018.

\end{thebibliography}
}

\begin{appendices}
\section{Data Collection}
\begin{figure}[!h]
\centering
\includegraphics[width=0.5\linewidth]{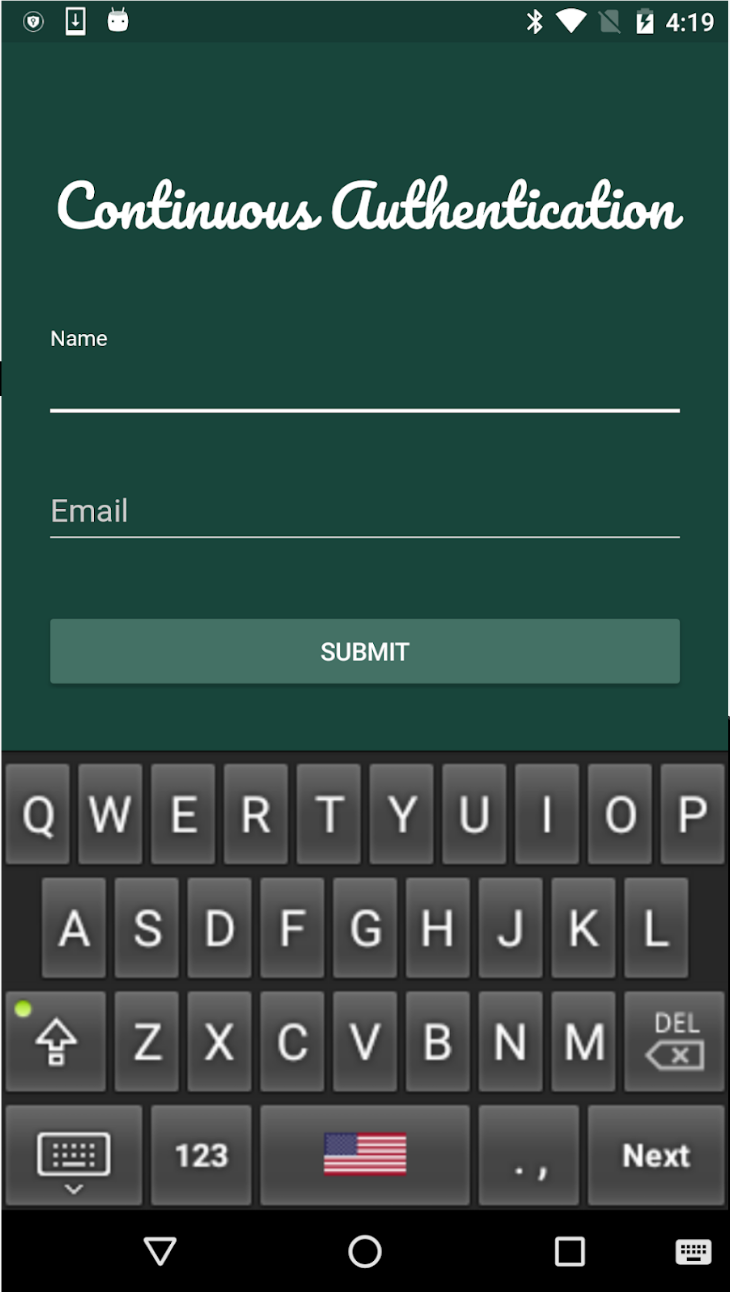}
\caption{Screenshot of our data collection app. The user downloads the Android app from Google Play Store and registers themselves. The app continuously runs in the background while passively acquiring sensor data.}
\label{fig:app}
\end{figure}

Screenshots of the Android application that was created for passively acquiring data from the sensors are shown in Figure~\ref{fig:app}. This application automatically turns on whenever the smartphone boots and continuously runs in the background while passively recording sensor data.
Data from the smartphone users were acquired on their own personal devices. The five most common Android smartphones used by the users in our dataset are given in Table~\ref{tab:phones}.
\begin{table}[!h] 
\footnotesize
\caption{Five most common Android smartphones used by the users in our dataset.}
\centering
\begin{tabular}{|l|c|c|}
\noalign{\hrule height 1.5pt}
\textbf{Model} & \textbf{Android Version} & \textbf{API version}\\
\noalign{\hrule height 1.0pt}
Google Pixel & 3.18.70 & 27\\ \hline
OnePlus&  3.18.70 & 26\\ \hline
Redmi Note 4 & 3.18.13 & 24\\ \hline
Nexus 5 & 3.04.00 & 23\\ \hline
MotoG3 & 3.10.49 & 23\\
\noalign{\hrule height 1.5pt}
\end{tabular}
\label{tab:phones}
\end{table}

\section{Modalities}
We list the 30 different modalities acquired in our dataset in Table~\ref{tab:all_modalities}.
\begin{table}[!h] 
\caption{Modalities acquired in our dataset. The eight modalities studied in this paper are highlighted in bold.}
\centering
\footnotesize
\resizebox{\linewidth}{!}{
\begin{tabular}{|l|l|}
\noalign{\hrule height 1.5pt}
\textbf{Accelerometer} & \textbf{Smartphone's acceleration in X, Y, Z plane}\\ \hline
 Application Usage & Name of the smartphone application used\\ \hline
Battery Levels & Percentage of battery charge left\\ \hline
Bluetooth Connections & Bluetooth connection names around device\\ \hline
Brightness Levels & Screen brightness level\\ \hline
Cell Tower Connections & Cell tower names around device\\ \hline
File Read In/Write Out & Files that were read from or written to device's disk\\ \hline
Glance Gesture & User glanced at their smartphone\\ \hline
\textbf{GPS Location} & \textbf{User's GPS location (latitude, longitude)}\\ \hline
\textbf{Gravity Sensor} & \textbf{Direction and magnitude of gravity}\\ \hline
\textbf{Gyroscope Gesture} & \textbf{Rate of rotation of the device in X, Y, and Z planes}\\ \hline
Heart Sensor & User's heart-rate beats-per-minute\\ \hline
Humidity Sensor  & Ambient air humidity percentage\\ \hline
\textbf{Keystroke Dynamics}    & \textbf{Key hold time, finger area and finger pressure}\\ \hline
Light Sensor & Ambient light measure\\ \hline
\textbf{Linear Acceleration} & \textbf{Linear acceleration in X, Y, and Z planes}\\ \hline
 \textbf{Magnetometer}   & \textbf{Earth's magnetic field in X, Y, and Z planes}\\ \hline
 NFC Connections & NFC connection names around device\\ \hline
 Orientation Sensor    & Device's orientation in X, Y, and Z planes\\ \hline
 Pickup Gesture & Device was picked up\\ \hline
 Barometer & Atmospheric pressure\\ \hline
 \textbf{Rotation Sensor} & \textbf{Device's rotation in X, Y, and Z planes}\\ \hline
 Screen Touch Sensor & Touch location\\ \hline
 Step Counter    & Number of steps walked\\ \hline
 Step Detector & User is walking\\ \hline
 Temperature Sensor & Ambient temperature\\ \hline
 Tilt Detector & Device is tilted\\ \hline
 Volume Levels & Level of volume set by user\\ \hline
 Wake up Gesture & Device is turned on\\ \hline
 WiFi Connections & WiFi connection names\\
\noalign{\hrule height 1.5pt}
\end{tabular}
}
\label{tab:all_modalities}
\end{table}

\section{Implementation Details}
\label{appndx:details}
The proposed Siamese LSTM is implemented in Keras 2.2.4\footnote{\url{https://keras.io}} with TensorFlow 1.9.0\footnote{\url{https://www.tensorflow.org/}} backend. The LSTM is implemented exactly as documented in the Keras website\footnote{\url{https://keras.io/layers/recurrent/}}.
\begin{table}[!h] 
\footnotesize
\caption{Training Details}
\centering
\begin{tabular}{|l|l|}
\noalign{\hrule height 1.5pt}
Batch Size & 82,240\\ \hline
Learning Rate & 0.05\\ \hline
Decay & 0.0\\ \hline
$\beta_1$ & 0.9\\ \hline
$\beta_1$ & 0.999\\ \hline
\noalign{\hrule height 1.5pt}
\end{tabular}
\label{tab:details}
\end{table}

\end{appendices}
\end{document}